\documentclass{sig-alternate}

\usepackage{graphicx}
\usepackage{epstopdf}

\usepackage{etoolbox}
\makeatletter
\def\@copyrightspace{\relax}
\makeatother

\begin{document}
%
\conferenceinfo{WOODSTOCK}{'97 El Paso, Texas USA}

\title{Sequential Feature Explanations for Anomaly Detection}


\author{
\alignauthor
Md Amran Siddiqui and Alan Fern and Thomas G. Dietterich and Weng-Keen Wong\\
       \affaddr{School of EECS}\\
       \affaddr{Oregon State University}\\
       \email{\{siddiqmd,afern,tgd,wong\}@eecs.oregonstate.edu}
}

\maketitle
\begin{abstract}
In many applications, an anomaly detection system presents the most anomalous data instance to a human analyst, who then must determine whether the instance is truly of interest (e.g. a threat in a security setting). Unfortunately, most anomaly detectors provide no explanation about why an instance was considered anomalous, leaving the analyst with no guidance about where to begin the investigation. To address this issue, we study the problems of computing and evaluating sequential feature explanations (SFEs) for anomaly detectors. An SFE of an anomaly is a sequence of features, which are presented to the analyst one at a time (in order) until the information contained in the highlighted features is enough for the analyst to make a confident judgement about the anomaly. Since analyst effort is related to the amount of information that they consider in an investigation, an explanation's quality is related to the number of features that must be revealed to attain confidence. One of our main contributions is to present a novel framework for large scale quantitative evaluations of SFEs, where the quality measure is based on analyst effort. To do this we construct anomaly detection benchmarks from real data sets along with artificial experts that can be simulated for evaluation. Our second contribution is to evaluate several novel explanation approaches within the framework and on traditional anomaly detection benchmarks, offering several insights into the approaches.
\end{abstract}




\section{Introduction}

Anomaly detection is the problem of identifying anomalies in a data set, where anomalies are those points that are generated by a process that is distinct from the process generating ``normal" points. Statistical anomaly detectors address this problem by seeking statistical outliers in the data. In most application, however, statistically outliers will not always correspond to the semantically-meaningful anomalies. For example, in a computer security application, a user may be considered statistically anomalous due to an unusually high amount of copying and printing activity, which in reality has a benign explanation and hence is not a true anomaly. Because of this gap between statistics and semantics, an analyst typically investigates the statistical outliers in order to decide which ones are likely to be true anomalies and deserve further action.

Given an outlier point, an analyst faces the problem of analyzing the data associated with that point in order to make a judgement about whether it is an anomaly or not. Even when points are described by just tens of features, this can be challenging, especially, when feature interactions are critical to the judgement. In practice, the situation is often much worse with points being described by thousands of features. In these cases, there is a significant risk that even when the anomaly detector passes a true anomaly to the analyst, the analyst will not recognize the key properties that make the point anomalous due to information overload. This means that, in effect, the missed anomaly rate of the overall system is a combination of the miss rates of both the anomaly detector and the analyst. Thus, one avenue for improving detection rates is to reduce the effort required by an analyst to correctly identify anomalies, with the intended side-effect of reducing the analyst miss rate.

In this paper, we consider reducing the analyst's detection effort by providing them with explanations about why points were judged to be anomalous by the detector. Given such an explanation, the analyst can minimize effort by focusing the investigation on information related to the explanation.

Our first contribution is to introduce an intuitive and simple form of explanation, which we refer to as \emph{sequential feature explanations (SFEs)}. Given a point judged to be an outlier by a detector, an SFE for that point is an ordered sequence of features, where the order indicates the importance with respect to causing a high outlier score. An SFE is presented to the analyst by incrementally revealing the features one at a time, in order, until the analyst has acquired enough information to make a decision about whether the point is an anomaly or not (e.g. in a security domian, threat or non-threat). The investigative work of the analyst is roughly related to the number of features that must be revealed. Hence, the goal for computing SFEs is to minimize the number of features that must be revealed in order for the analyst to confidently identify true anomalies.

Our second contribution is to formulate a quantitative evaluation methodology for evaluating SFEs, allowing for the comparison of different SFE algorithms. The key idea of the approach is to construct a simulated analyst for each anomaly detection benchmark using supervised learning and ground truth about which points are anomalies. The simulated analyst can then be used to evaluate the quality of SFEs with respect to the number of features that must be revealed to reach a specified confidence level. To the best of our knowledge this is the first methodology for quantitatively evaluating any type of anomaly explanation method.

Our third contribution is to define several algorithms for computing SFEs that can be applied to any density-based anomaly detector. The main requirement of the algorithms is that it is possible to (approximately) compute joint marginals of a detector's density function, which is an operation that is supported for most commonly-used densities.

Finally, our fourth contribution is to provide an empirical investigation of several methods for computing SFEs. Our primary evaluations use a recently constructed set of anomaly detection benchmarks derived from real-world supervised learning data. In addition we provide an evaluation on the standard KDD-Cup benchmark. The investigation leads to a recommended method and additional insights into the methods.

The remainder of the paper is organized as follows. Section \ref{sec:related} reviews related work on explanations for both supervised learning and anomaly detection. Next, Section \ref{sec:anomaly} presents the anomaly-concepts formulation used in this paper. Section \ref{sec:SFE} then more formally presents the concept of SFEs and possible quality metrics. Section \ref{sec:methods} describes and contrast several methods for computing SFEs. Section \ref{sec:framework} then introduces our quantitative evaluation framework for SFEs and finally Section \ref{sec:experiments} presents experiments evaluating the introduced methods within the framework.

\section{Related Work}
\label{sec:related}

The problem of computing explanations for both supervised learning and unsupervised settings, such as anomaly detection, has received relatively little attention. Related work in the area of supervised classification aims to provide explanations about why a classifier predicted a particular label for a particular instance. For example, a number of methods have been proposed to produce explanations in the form of relevance scores for each feature, which indicate the relative importance of a feature to the classification decision. Such scores have been computed by comparing the difference between a classifier's prediction score and the score when a feature is assumed to be unobserved \cite{robnik2008explaining}, or by considering the local gradient of the classifier's prediction score with respect to the features for a particular instance \cite{baehrens2010explain}.

Other work has considered how to score features in a way that takes into account the joint influence of feature subsets on the classification score, which usually requires approximations due to the exponential number of such subsets \cite{strumbelj2010efficient,vstrumbelj2013explaining}. Since these methods are typically based on the availability of a class-conditional probability function, they are not directly generalizable to computing explanations for anomaly detectors. Our experiments, however, do evaluate a method, called Dropout, which is inspired by the approach of \cite{robnik2008explaining}.

The form of such feature-relevance explanations is similar in nature to our SFEs in that they provide an ordering on features. However, prior work has not explicitly considered the concept of sequentially revealing features to an analyst, which is a key part of the SFE proposal for reducing analyst effort.


Prior work on feature-based explanations for anomaly detection has focused primarily on computing explanations in the form of feature subsets. Such explanations are intended to specify the subset of features that are jointly responsible for an object receiving a high anomaly score. For example, Micenkova, et al. \cite{micenkova2013explaining} computed a subset of features such that the projection of the anomalous object onto the features shows the greatest deviation from normal instances. One issue with this approach is that the computation of an explanation is independent of the anomaly detector being employed. This is contrary to the goal of trying to explain why a particular anomaly detector judged a particular object to be anomalous. In contrast, the explanation approaches we consider in this paper are sensitive to the particular anomaly detector.

Other work on computing feature-subset explanations \cite{dang2013local} developed an anomaly detection system called LODI which includes a specialized explanation mechanism for the particular anomaly detector. A similar approach is considered by Dang, et al. \cite{dang2014discriminative}, where the anomaly detection mechanism directly searches for discriminative subspaces that can be used for the purpose of explanation. In contrast, the explanation approaches we consider in this work can be instantiated for any anomaly detection scheme based on density estimation, which includes a large fraction of existing detectors.


Existing approaches for evaluating explanations methods in both supervised and unsupervised settings are typically quite limited in their scope. Often evaluations are limited to visualizations or illustrations of several example explanations \cite{baehrens2010explain,dang2014discriminative} or to testing whether a computed explanation collectively conforms to some known concept in the data set \cite{baehrens2010explain}, often for synthetically generated data. Prior work has not yet proposed a larger scale quantitative evaluation methodology for explanations, which is one of the main contributions of our work.

\section{Anomaly Detection Formulation}
\label{sec:anomaly}

We consider anomaly detection problems defined over a set of $N$ data points $\{x_1,\ldots,x_N\}$, where each point $x_i$ is an $n$ dimensional real-valued vector. The set contains a mixture of \emph{normal points} and \emph{anomaly points}, where generally the normal points account for an overwhelming fraction of the data. In most applications of anomaly detection, the anomaly points are generated by a distinct process from that of the normal points, in particular, a process that is important to detect for the particular application. For example, the data points may describe the usage behavior of all users of a corporate computer network and the anomalies may correspond to insider threats.

Since $N$ is typically large, manual search for anomalies through all points is generally not practical. Statistical anomaly detectors address this issue by seeking to identify anomalies by finding statistical outliers. The problem, however, is that not all outliers correspond to anomalies, and in practice an analyst must examine the outliers to decide which ones are likely to be anomalies. We say that an analyst \emph{detects} an anomaly when he or she is presented with an anomaly point and is able to determine that there is enough evidence that the point is indeed an anomaly. The success of this approach depends on the anomaly detector's precision of identifying anomalies as outliers, and also on the analysts' ability to correctly detect anomalies. Without further assistance, an analyst may need to consider information related to all $n$ features of an anomaly point during analysis. In many cases, considering this information thoroughly will be impossible, increasing the chance of not detecting anomalies, which can be costly in many domains.

\section{Sequential Feature Explanations}
\label{sec:SFE}

In order to reduce the analyst's effort toward detecting anomalies, we propose to provide the analyst with \emph{sequential feature explanations (SFEs)} that attempt to efficiently explain why a point was considered to be an outlier. A length $k$ SFE for a point is an ordered list of feature indices $E=(e_1,\ldots,e_k)$, where $e_i\in \{1,\ldots,n\}$. The intention is that features that appear earlier in the order are considered to be more important to the high outlier score of a point (e.g. $x_{e_1}$ is the most important). We will use the notation $E_i$ to denote the set of the first $i$ feature indices of $E$. Also, for any set of feature indices $S$ and a data point $x$, we let $x_S$ denote the projection of $x$ onto the subspace specified by $S$.

Given an SFE $E$ for a point $x$, the point is incrementally presented to the analyst by first presenting only feature $x_{E_1}$. If the analyst is able to make a judgement based on only that information then we are finished with the point. Otherwise, the next feature is added to the information given to the analyst, that is, the analyst now sees $x_{E_{2}}$. The process of incrementally adding features to the set of presented information continues until the analyst is able to make a decision. The process may also terminate early because of time constraints; however, we don't study that case in this paper.

For normal points, the incremental presentation of SFEs may not help the analyst more efficiently exonerate the points. In contrast, for anomalies, it is reasonable to expect that an analyst would be able to detect the anomalies by considering a much smaller amount of information than without the SFE, which should reduce the chance of missed detections. We assume that the amount of analyst effort is a monotonically increasing function of the number of features considered. This motivates measuring the quality of an SFE for a target by the number of features that must be revealed to an analyst for correct detection. More formally, given an anomaly point $x$, an analyst $a$, and an SFE $E$ for $x$, the \emph{minimum feature prefix}, denoted $\mbox{MFP}(x,a,E)$, is the minimum number of features that must be revealed to $a$, in the order specified by $E$, for $a$ to detect $x$ as an anomaly.

While MFP provides a quantitative measure of SFE quality, its definition requires access to an analyst. This complicates the comparison of SFE computation methods in terms of MFP. Section \ref{sec:framework} addresses this issue and describes an approach for conducting wide evaluations in terms of MFP.




\section{Explanation Methods}
\label{sec:methods}

We now consider methods for computing SFEs for anomaly detectors. Prior work on computing explanations for anomaly detectors has either computed explanations that do not depend on the particular anomaly detector used (e.g. \cite{micenkova2013explaining}) or used methods that were specific to a particular anomaly detector (e.g. \cite{dang2013local}). We wish to avoid the former approach, since intuitively an explanation should attempt to indicate why the particular detector being employed found a point to be an outlier. Considering the latter approach, we seek more general methods that can be applied more widely across different detectors. Thus, here we consider explanation methods for the widely-studied class of \emph{density-based detectors}.\footnote{Our methods can actually be employed on the more general class of ``score-based detectors" provided that scores can be computed given any subset of features. For simplicity, we focus on density-based detectors in this paper, where the density function is used to compute scores.}

Density-based detectors operate by estimating a probability density function $f(x)$ (e.g. a Gausssian mixture) over the entire set of $N$ points and treating $f$ as the density over normal points. This is reasonable under the usual assumption that anomalies are very rare compared to normals. Points are then ranked according to ascending values of $f(X)$ so that the least normal objects according to $f$ are highest in the order. Our methods do not assume knowledge of the form of $f$, but do require an interface to $f$ that allows for joint marginal values to be computed. That is, for any subset of feature indices $S$ and point $x$, we require that we can compute $f(x_S)$. For many choices of $f$, such as mixtures of Gaussians, these joint marginals have simple closed forms. If no closed form is available, then exact or approximate inference techniques (e.g., MCMC) may be employed.

It is worth noting that by considering SFE methods that depend on the anomaly detector being used, the performance in terms of MFP will depend on the quality of the anomaly detector as well as the SFE method. For example, consider a situation where the anomaly detector judges an anomaly point $x$ to be an outlier for reasons that are not semantically relevant to why $x$ is an anomaly. The SFE for $x$ is not likely to help the analyst to more efficiently determine that $x$ is an anomaly, since the semantically critical features may appear late in the ordering. While this is a possibility, it is out of control of the SFE method. Thus, when designing SFE methods we will assume that outlier judgements made by $f$ are semantically meaningful with respect to the application. We now present our two main classes of SFE methods which we refer to as \emph{marginal methods} and \emph{dropout methods}.

\subsection{Marginal Methods}

Here we consider modeling the analyst as a Bayesian classifier that assumes normal points are generated according to $f$ and that anomalies have a uniform distribution $u$ over the support of the feature space, a reasonable assumption in the absence of prior knowledge about the anomaly distribution. Given a point $x$, an SFE $E$, and a number of revealed features $i$, such an analyst would make the decision of whether $x$ is an anomaly or not by comparing the likelihood ratio $\frac{f(x_{E_i})}{u(x_{E_i})}$ to some threshold. Since $u$ is assumed to be uniform, this is equivalent to comparing the joint marginal $f(x_{E_i})$ to a threshold. Intuitively this means that if our goal is to cause the analyst to quickly decide that $x$ is an anomaly, we should chose an $E$ that yields small values of $f(x_{E_i})$, particularly for small $i$.

This leads to our first SFE method, called \emph{sequential marginal (SeqMarg)}. The SeqMarg method adds one feature to the SFE $E=(e_1,\ldots,e_k)$  at a time, at each step adding the feature that minimizes the joint marginal density with the previously-selected features. More formally, SeqM computes the following explanation:
$$\mbox{\bf SeqMarg:          } e_i = \arg\min_{j \in \overline{E}_{i-1}} f(x_{E_{i-1}},x_j)$$
where $\overline{S}$ is the complement of set $S$. SeqM requires $O(kn)$ joint marginal computations in order to compute an explanation of length $k$. Note that due to the inherent greediness of SeqM, $x_{E_i}$ may not necessarily be the optimal set of $i$ features for minimizing $f$. Rather, if the goal were to optimize for a particular value of $i$, we would need to consider all $O(n^i)$ feature subsets of size $i$. However, our problem formulation does not provide us with a target value of $i$, and thus SeqM offers a more tractable approach that focuses on minimizing $f$ as quickly as possible in a greedy manner.

In addition to SeqMarg we also consider a computationally cheaper alternative, called \emph{independent marginal (IndMarg)}, which only requires the computation of individual marginals $f(x_i)$. This approach simply selects an explanation $E$ for $x$ by sorting the features in increasing order of $f(x_i)$. This only requires $O(n)$  marginal computations for computing an explanation of any length. IndMarg offers a computationally cheaper alternative to SeqMarg, but fails to capture joint feature interactions. For example, SeqMarg will select $e_i$ in a way that optimizes the joint value when combined with previous features $E_{i-1}$. Instead, IndMarg ignores interactions with previously-selected features. Thus, IndMarg serves as a baseline for understanding the importance of accounting for joint feature interactions when computing explanations.

\subsection{Dropout Methods}

The next two methods are inspired by the work of Robnik-Sikonja and Kononenko \cite{robnik2008explaining} on computing feature-relevance explanations for supervised classifiers. In their work, the relevance score for a feature is the difference between the classification score when the feature is provided to the classifier and the classification when the feature is omitted (``dropped out''). The analogous approach for anomaly detection is to score features according to the change in the density value when the feature is included and when the feature is not included, or marginalized out. This yields the first dropout method, referred to as \emph{independent dropout (IndDO)}: given a point $x$, each feature is assigned a score of $f(x-x_i)-f(x)$, where we abuse notation and denote the removal of $x_i$ from $x$ by $x-x_i$. Intuitively, features with larger scores are ones that make the point appear most normal when removed. The SFE $E$ is then obtained by sorting features in decreasing order of score.

We can also define a sequential version of dropout, by following the same recipe we considered for IndMarg versus SeqMarg. Let the \emph{sequential dropout (SeqDO)} be defined as follows:
$$\mbox{\bf SeqDO:          } e_i = \arg\max_{j\in \overline{E}_{1:i-1}} f(x_{\overline{E}_{i}}-x_j).$$
This approach requires the same number of marginal computations as SeqMarg. This algorithm can be viewed as a dual of SeqMarg in that it measures the contribution of feature sets according to how much more normal a point looks after their removal, whereas SeqMarg measures how abnormal a point looks with only those features included.

\section{Framework for Evaluating Explanations}
\label{sec:framework}

%
%
%
%

There are at least two challenges involved in evaluating anomaly-explanation methods. First, compared to supervised learning, the area of anomaly detection has many fewer established benchmark data sets, particularly benchmarks based on real-world data. Second, given a benchmark data set, it is not immediately clear how to quantitatively evaluate explanations, since the benchmarks do not come with either ground truth explanations or analysts.

Here we describe an evaluation framework that addresses both issues. We address the first issue by drawing on recent work on constructing large numbers of anomaly detection benchmarks based on real-world data. We address the second issue by using supervised learning to construct a simulated analyst that can be applied to quantitatively evaluate our explanations in terms of MFP. Below we expand on both of these points.

\subsection{Anomaly Detection Benchmarks}

Recent work \cite{emmott2013systematic} described a methodology for systematically creating anomaly detection benchmarks from supervised learning benchmarks (either classification or regression). Given the huge number of real-world supervised learning benchmarks, this allows for a corresponding huge number and diverse set of anomaly detection benchmarks. Further, these benchmarks can be created to have controllable and measurable properties, such as anomaly frequency and ``clusteredness" of the normal and anomalous points. We briefly sketch the main idea. Given a supervised classification data set, called the \emph{mother set}, the approach selects one or more of the classes to represent the anomaly class, with different choices giving rise to different properties of the anomaly class. The union of the other classes represents the normal class. Individual anomaly detection benchmarks are then created by sampling the normal and anomaly points at specified proportions.

Table \ref{datasetSummary} gives a summary of the benchmarks from Emmott et al. \cite{emmott2013systematic} used in our experiments. For example, the UCI data set shuttle was used as a mother set to create 1600 distinct anomaly detection benchmarks. The number of points in the shuttle benchmarks range from 3570 to 9847. The number of anomalies ranges from 8 to 984. 

\subsection{Simulated Analyst}

\begin{figure*}[t]

\includegraphics[scale=0.6]{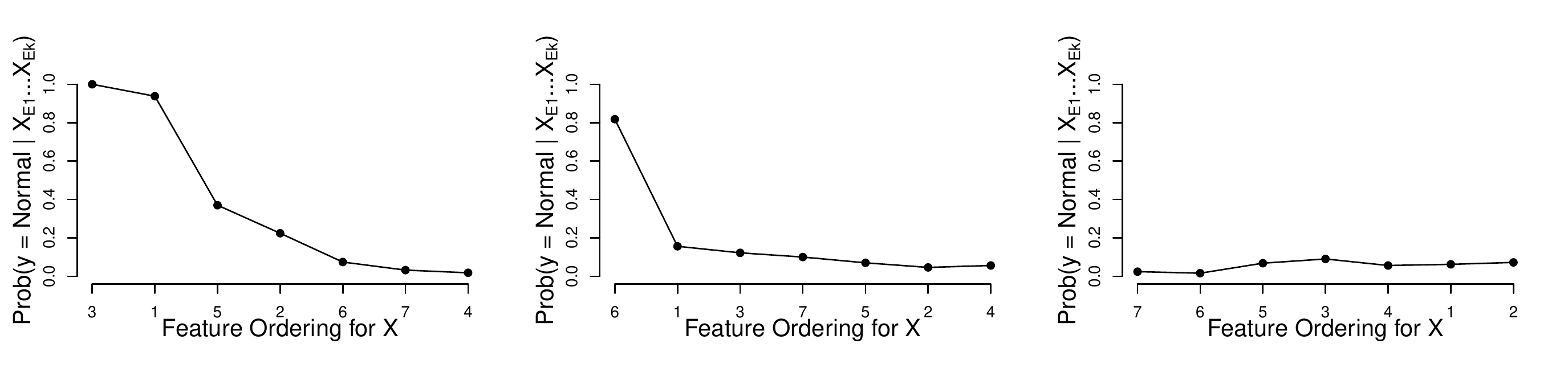}
\caption{Analyst Certainity Curves. These are example curves generated using our simulated analyst on anomalies from the Abalone benchmark using SFEs produced by SeqMarg. The x-axis shows the index of the feature revealed at each step and the y-axis shows the analyst certainty about the anomalies being normal. The leftmost curve shows an example of where the analyst gradually becomes certain that the point is anomalous, while the middle curve shows more rapidly growing certainty. The middle curves is an example of where the analyst is certain of the anomaly after the first feature is revealed and remains certain.}\label{fig:analystCurve}
\end{figure*}

We consider modeling an analyst as a conditional distribution of the normal class given a subset of features from a data point. More formally we model the analyst as a function $A(x,S)=P(\mbox{normal}\;|\; x_S)$, which returns the probability that point $x$ is normal considering only the features specified by the set $S$. We describe how we obtain this function in our experiments below. Given this function, a point $x$, and an SFE $E$ for $x$, we can generate an \emph{analyst certainty curve} that plots the analyst's certainty after revealing $i$ features, that is, $A(x,E_i)$ versus $i$. Figure \ref{fig:analystCurve} shows an example of three analyst curves from our experiments using our simulated analysts on a benchmark computed from the UCI Abalone dataset. The curves each correspond to a different anomaly in the data set using explanations computed using SeqMarg. We see that the different anomalies lead to different rates at which the analyst becomes certain of the anomaly, that is, certain that the point is not normal.

Recall that our proposed quality metric $\mbox{MFP}(x,a,E)$ measures the number of features that must be revealed to analyst $a$ according to SFE $E$ in order for $a$ to detect an anomaly $x$. Evaluating this metric requires that we define the conditions under which the analyst detects $x$. We model this by associating an analyst with a detection threshold $\tau \in [0,0.5]$ and saying that a detection occurs if $A(x,E_i) \leq \tau$, that is, the probability of normality becomes small enough. We will denote this analyst by $a(\tau)$. Given an $a(\tau)$ we can then compute the MFP for any anomaly point by recording number of features required for the analyst certainty curve to first drop below $\tau$.

Of course, there is no apriori basis for selecting a value of $\tau$. Thus, in our experiments, we consider a discrete distribution over values for $\tau$, $P(\tau)$, which models a range of reasonable thresholds. Given this distribution, we report the expected MFP---the expected value of $\mbox{MFP}(x,a(\tau),E)$---as the quantitative measure of SFE $E$ for anomaly $x$. In our experiments we define $P(\tau)$ to be uniform over the values 0.1, 0.2, and 0.3, noting that our results are consistent across a variety of reasonable choices for this distribution.

It remains to specify how we obtain the analyst function $A(x,S)$. Since our anomaly detection benchmarks are each derived from a mother classification data set, we can construct a training set over those points for the anomaly and normal classes. Given this training set, one approach to obtaining the analyst would be to learn a generative model, or joint distribution $P(\mbox{normal},x)$, which could be used to compute $A(x,S)$ by marginalizing out features not included in $s$. However, such generative models tend to be much less accurate in practice compared to discriminative models. However, learning a discriminative model $P(\mbox{normal}\;|\;x)$ does not directly support computing the probability for arbitrary subsets of $x$ as we require. While heuristics have been proposed for this purpose (e.g. Robnik-Sikonja and Kononenko \cite{robnik2008explaining}) we have found them to be unreliable when applied widely. Thus, in this work we follow a brute force approach. We simply pre-learn an individual discriminative model for each possible subset of features up to a maximum size $k$. Evaluating $A(x,S)$ then simply requires evaluating the model associated with the subset $S$.

When the number of features or number of data points is very large, it may not be possible to pre-learn all possible subsets. In such cases, one option is to learn and cache models on the fly as they are needed during evaluation (each model would be learned only once). We used this approach for the KDD-Cup results reported in our experiments.

\section{Empirical Evaluation}
\label{sec:experiments}

We now present our empirical evaluation on anomaly detection benchmarks from Emmott et al.~\cite{emmott2013systematic} and the commonly used KDDCup anomaly detection benchmark.

\subsection{Anomaly Detector}

For all of our experiments, we have chosen to use the Ensemble Gaussian Mixture Model (EGMM) as the anomaly detector. This detector was first described in Emmott et al. \cite{emmott2013systematic} and was shown to be a competitive density-based approach across a wide range of benchmarks. EGMM is based on learning a density function $f(x)$ represented as an ensemble of Gaussian mixture models (GMMs). The approach independently learns $M$ GMM models by training each one using the Expectation-Maximization (EM) procedure on bootstrap replicates of the data set. Then it discards the low-likelihood GMMs (if any) and retains others based on a pre-specified threshold. The number of components of the GMMs is varied across the ensemble. In our experiments, the ensembles included 45 GMMs, 15 each using 3, 4, and 5 components. The final EGMM density $f(x)$ is simply a uniform mixture of the densities of the retained GMMs. The EGMM approach addresses at least two pitfalls of using single GMM models. First, EM training can sometimes produce poor models due to bad local optima. Second, it is difficult to select the best number of components to use for a single model. EGMM gains robustness by performing model averaging over the variations.

One advantage of using the EGMM model is that it is straightforward to derive closed forms for the marginal density computations required by our explanation methods. In particular, the overall EGMM density $f$ can be viewed as a single large GMM model containing a mixture of all components across the ensemble. Since individual Gaussians have simple closed forms for marginal densities\cite{bishop2006pattern}, we can easily obtain closed forms for the mixture. It is worth noting that closed forms can also be derived for EGMM marginals when the data points are transformed by linear projections to reduce dimensionality (e.g. principle component analysis).

\subsection{Simulated Expert}

Recall that our evaluation framework is based on using supervised learning in order to obtain a simulated analyst. Our experiments are based on using Regularized Random Forests (RRFs) \cite{Deng2013} as the analyst model. The RRF model was selected for two primary reasons. First, RRFs are well-known to provide high accuracies that are competitive with the state-of-the-art across a wide range of classification problems. Second, RRFs are relatively efficient to train, which is important to our study, since we must train one RRF for each possible subset of features (up to some maximum size). We trained RRFs composed of 500 trees using 10-fold cross-validation in order to tune the RRF regularization parameters.

It is worth noting that our evaluation framework is potentially sensitive to the choice of analyst model, since different models will have different biases. It was beyond the scope of this first study to replicate all experiments using a qualitatively different model. This will be a point of future work.

\subsection{Evaluation on Benchmark Data Sets}

\begin{table*}[t]
\caption{Summary of the benchmark datasets}\label{datasetSummary}
\begin{center}
\begin{tabular}{| p{20mm} | p{16mm} | p{14mm} | p{20mm} | p{20mm} | p{20mm} |}
\hline
Mother Set 		&Original Problem Type 		&\#Features	& \# of Anomaly Benchmarks		&\# of Points per Benchmark (range) & \# of Anomaly per Benchark (range) \\
\hline
\hline
magic.gamma		&Binary 	&	10	& 1600 & 600 - 6180 & 5 - 618\\
\hline
skin		&Binary 		& 3	& 1200 & 10 - 9323 & 1 - 932\\
\hline
shuttle		& Multiclass	& 9	& 1600 & 3570 - 9847 & 8 - 984\\
\hline
yeast		& Multiclass	& 8	& 1600 & 70 - 1000 & 1 - 97\\
\hline
abalone		& Regression	& 7	& 1600 & 580 - 2095 & 1 - 209\\
\hline
concrete &	Regression & 	8	& 1200 & 190 - 1000 & 1 - 51\\
\hline
wine	&	Regression &	11	& 1600 & 2590 - 4112 & 3 - 411\\
\hline
\end{tabular}
\end{center}
\end{table*}

We run our evaluation on anomaly detection benchmarks, from Emmott et al.~\cite{emmott2013systematic}, derived from seven UCI mother sets. A summary of the benchmarks are given in Table \ref{datasetSummary}. There are over 10,000 benchmarks in total, which contain a number of points ranging from 10 to 9800 and a number of anomalies ranging from 1 to 930. An EGMM model was fit for each of the benchmarks to serve as the anomaly detector, and RRF models were trained for each mother set on all possible feature subsets. For this first study, we have chosen to focus on benchmarks with relatively small dimensionality in order to allow for a large scale study, which requires training large numbers of EGMM models (over 10,000) and RRF analyst models. All data from these experiments, including the analysts models, will be made publicly available.

\begin{figure*}[t]
\begin{center}
\includegraphics[bb = 55 315 560 475,clip=true,width=.99\textwidth]{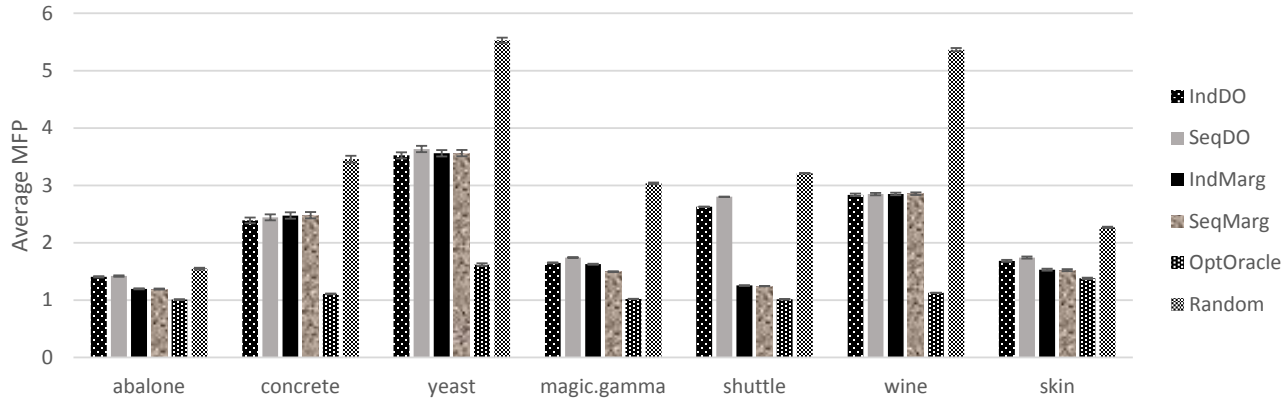}
\end{center}
\caption{Performance of explanation methods on benchmarks. Each group of bars shows the performance of the six methods on benchmarks derived from a single mother set. The bars show the expected MFP averaged across anomalies in benchmarks for the corresponding mother set. 95\% confidence intervals are also shown.}\label{barChartExplMethodsEGMM}
\end{figure*}

\begin{figure*}
\begin{center}
\includegraphics[bb = 55 315 560 475,clip=true,width=.99\textwidth]{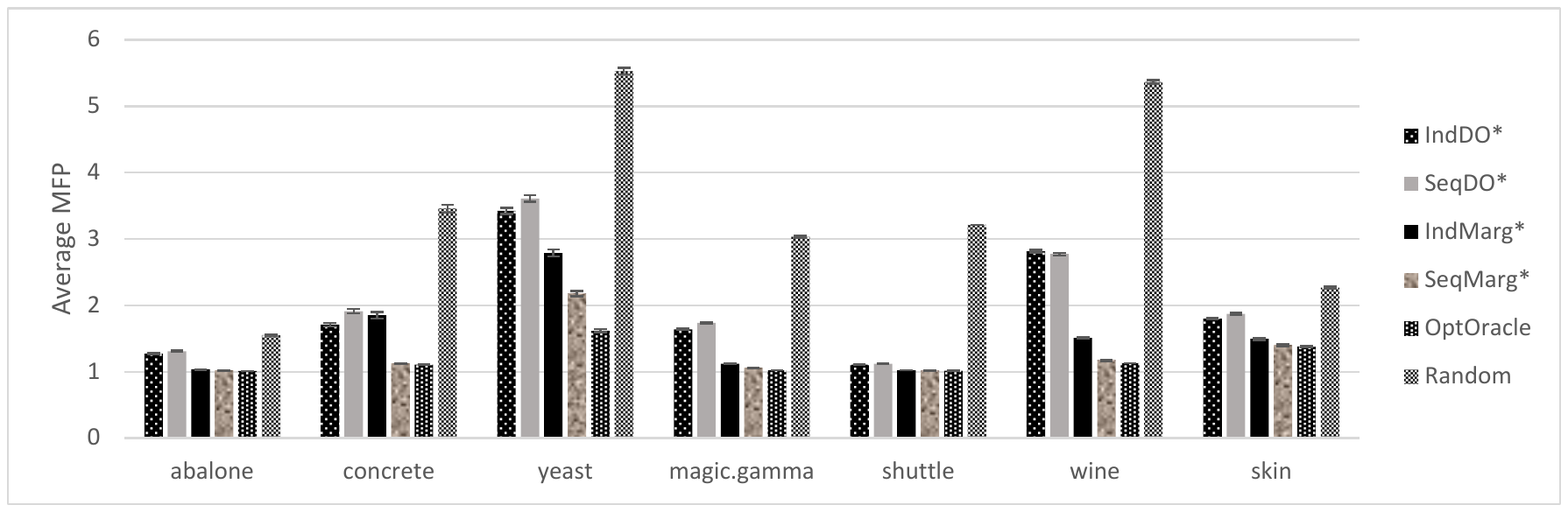}
\end{center}
\caption{Performance of explanation methods on benchmarks when using an oracle anomaly detector. Each group of bars shows the performance of the six methods on benchmarks derived from a single mother set. The bars show the expected MFP averaged across anomalies in benchmarks for the corresponding mother set. 95\% confidence intervals are also shown.} \label{barChartExplMethodsOracle}
\end{figure*}


We evaluated six methods for computing SFEs. These included the four methods from Section \ref{sec:methods}: SeqMarg, IndMarg, SeqDO, and IndDO. In addition, we evaluated a random explanation method. In the case of random, we report the average performance across 100 randomly generated SFEs. 

Finally, in order to provide an lower-bound on attainable performance (lower MFP is better) we consider an optimal oracle method, \emph{OptOracle}. This method is allowed access to the simulated analyst and for each number of features $i$ computes the optimal feature subset of size $i$. More formally, for each value of $i$, OptOracle finds the feature subset $S_i$ that minimizes the analyst's conditional probability $P(\mbox{normal}\;|\; x_{S_i})$.  The MFP achieved by OptOracle for an anomaly $x$, given a particular analyst threshold $\tau$ (recall Section \ref{sec:framework}), is the minimum value of $i$ such that $P(\mbox{normal}\;|\; x_{S_i})<\tau$. Note that OptOracle is not constrained to produce ``sequential explanations"---rather, \mbox{OptOracle} can produce an $S_i$ that does not necessarily contain $S_{i-1}$. This gives OptOracle an additional advantage compared to the other methods which are constrainted to produce SFEs. Clearly, OptOracle represents an upper bound on the performance of any SFE method that is evaluated with respect to the simulated analyst.

For each of the 10,000 benchmarks, we used the corresponding EGMM model to rank the points. For the anomaly points ranked in the top 10\%, we computed SFEs using each of the six methods. This choice is an attempt to model the fact that, in actual operation, only highly ranked anomalies will be presented to the expert.
The expected MFP was computed for each SFE using a distribution over analyst thresholds that was uniform over the values 0.1, 0.2, and 0.3. For each mother set, we then report the average MFP across the anomalies derived from that mother set. These average MFPs are shown in Figure \ref{barChartExplMethodsEGMM} along with 95\% confidence intervals.

We first note that our observations below are not sensitive to the choice of focusing on anomalies in the top 10\%. Indeed, we have also compiled results for other percentage points, including using all anomalies. The main observations are qualitatively similar across all of these choices.



\textbf{Comparison to Random and OptOracle.} We observe in Figure \ref{barChartExplMethodsEGMM} that all of the SFE methods outperform random explanations and often do so by a large margin. Comparing to OptOracle we see that, for three benchmarks---{\sc concrete}, {\sc yeast}, and {\sc wine}---the lower bound provided by OptOracle is significantly better than our best SFE method. This gap could be due to either: 1) suboptimal SFE computations, 2) a poor match between the anomaly detector's notion of outlier versus the analyst's notion of anomaly, or 3) the fact that OptOracle is not constrained to output sequential explanations. We will investigate this further below. 

For the remaining four mother sets, we see that the marginal methods are quite close to the lower bound of OptOracle, though there is still some room for improvement. Finally, it is worth noting that OptOracle is able to achieve MFPs of close to 1 for most of the mother sets. Thus, on average, for these data sets, a single feature is sufficient to allow for correct analyst detections.
 
\textbf{Independent versus Sequential.} It is reasonable to expect that the sequential version of the marginal and dropout methods will outperform the independent versions. This is because the sequential versions attempt to account more aggressively for feature interaction when computing SFEs, which requires additional computation time. However, we see that overall there is very little difference in performance between the independent and sequential methods. That is, SeqMarg and IndMarg (as well as SeqDO and IndDO) achieve nearly identical performance. The only exception is in magic.gamma where there a small, but statistically significant, advantage (according to a paired t-test) of SeqMarg over IndMarg. One possible explanation for these results is that feature interactions are not critical in these domains for detecting anomalies. This explanation is supported by the fact that OptOracle is able to achieve average MFPs close to one.

\textbf{Marginal versus Dropout.} Recall that the marginal and dropout methods are dual approaches. Marginal evaluates a set of features in terms of how abnormal those features alone make a point appear, while dropout evaluates a set by the increase in normality score when the features are removed. We see that overall the marginal methods are never significantly worse than dropout and significantly better on {\sc abalone}, {\sc magic.gamma}, {\sc shuttle}, and {\sc skin}. The difference is particularly large on {\sc shuttle}, where the marginal methods are close to OptOracle and the dropout methods are closer to random.

One possible explanation is that we have observed that often dropout will produce a ``weaker signal" compared to marginal when making early decisions. For example, when considering single features, the differences in scores produced by dropout for those features are often much smaller than the differences produced by marginal. This can make dropout less robust for early decisions, which are the most important ones for achieving small MFP scores. Recall, that the dropout method was inspired by prior work on explanations for supervised learning. The results here suggest that it is worth investigating adaptations of marginal to the supervised setting.

\subsection{Comparing Methods with Oracle Detectors}

Since the SFE methods make their decisions based on the anomaly detector's density function $f$, the results above reflect both the SFE methods and the quality of the detector. Here we attempt to factor out the performance of the SFE methods themselves by supplying the methods with an oracle anomaly detector. To do this we simply replace the use of $f$ with the simulated analyst's conditional probability function $P(\mbox{normal}\;|\; x_S)$, which we can compute for any feature subset $S$. For example, the first feature selected by SeqMarg is the $x_i$ that minimizes $P(\mbox{normal}\;|\; x_i)$. Note that this is also the first feature that would be selected by OptOracle. Unlike OptOracle, however, SeqMarg is sequentially constrained and will select the second feature as the one that works best when combined with the first selected feature.

Figure \ref{barChartExplMethodsOracle} shows results for all methods using the oracle detectors. We use a `*' to indicate that a method is using an oracle detector, for example, SeqMarg* is the oracle version of SeqMarg.

\textbf{Comparison to OptOracle.} The primary observation is that SeqMarg* performs nearly identically to OptOracle in all but one domain. Any difference between SeqMarg* and OptOracle would reflect the loss in performance due to requiring sequential explanations. For these data sets, there is little to no loss. This is good news, since the motivation for considering sequential explanations is to reduce the analyst's effort. In particular, the sequential constraint means that the analyst is shown an incrementally growing set of information. Rather, without the constraint, OptOracle could potentially show completely different sets of features from step to step, which is arguably less desirable from a usability perspective.

\textbf{Independent versus Sequential.} Here, we see that SeqMarg* is often outperforming IndMarg* and sometimes by significant amounts. This is in contrast to the results obtained when using EGMM as the anomaly detector. This observation indicates that reasoning about feature interactions, as done by SeqMarg*, can be important with higher quality anomaly detection models. This leaves an open question of whether we will be able to observe this advantage when using non-oracle anomaly detection models on realistic benchmarks.

\textbf{Dropout versus Marginal.} The marginal methods show consistently better performance when using oracle detectors. The performance gap is quite large in several of the benchmarks. This provides evidence that the marginal approach is generally a better way of computing SFEs. Again we hypothesize that this is due to the ``weak signal" during early decisions observed for the dropout method. 



\subsection{Evaluation on KDDCup'99 Data set}

We now show results on the UCI KDDCup intrusion detection bechmark \cite{hettich1999uci}. The points in this data set have 41 features, and we consider a subset of the data containing instances involving http service. The resulting benchmark contains approximately 620K points with approximately 4K anomaly points representing network intrusions. We again employed EGMM as the anomaly detector. It was infeasible to train a simulated analyst on all feature subsets, thus we followed the adaptive approach described Section \ref{sec:framework} where only the subset of models required during the evaluation process was learned and cached. Overall this resulted in approximately 7.5K RFF models being trained. In this domain, the EGMM model was quite effective and ranked all anomalies very close to the top of the ranked list. Thus, we evaluate on all anomalies in this domain.

\begin{figure}[t]
\begin{center}
\includegraphics[bb = 55 285 560 505,clip=true,width=.47\textwidth]{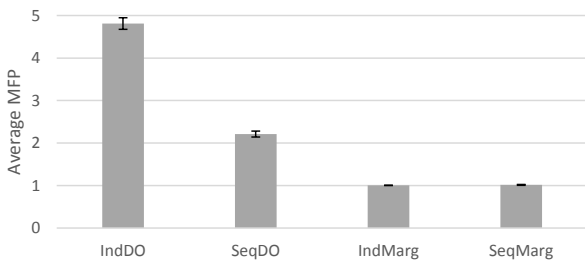}
\end{center}
\caption{Performance of different explanation methods on the KDDCup benchmark. 95\% confidence intervals are also shown.}\label{barChartKDDCup99}
\end{figure}

Figure \ref{barChartKDDCup99} shows the average MFP achieved by our methods. It is clear that the marginal methods are significantly better than the dropout methods here. In particular, both SeqMarg and IndMarg achieve an average MFP close to one, which is the smallest possible. This indicates that the combination of EGMM and marginal explanations is very effective in this domain. In particular, the simulated analyst only needed to be shown a single feature on average in order to correctly detect the anomalies.

We again hypothesize that the much weaker performance of the dropout methods is due to the ``weak signal" they provide for early decisions. This problem is only amplified in the context of larger numbers of features, as is the case for the KDDCup data.

\section{Main Observations}

The main observations from the above experiments can be summarized as follows.
\begin{itemize}
\item All of the introduced SFE methods significantly outperformed randomly generated SFEs.
\item The marginal methods were generally no worse and sometime significantly better than the dropout methods. 
\item When using the EGMM anomaly detector, we observed little to no difference between the performance of sequential versus independent methods. 
\item When using the oracle anomaly detector, SeqMarg significantly outperformed IndMarg, which suggests that in general sequential methods can outperform independent methods.
\item Overall, based on our results, SeqMarg is the recommended method for computing SFEs, among the methods we studied.
\end{itemize}

\section{Summary}

This paper introduced the concept of sequential feature explanations (SFEs) for anomaly detection. The main motivation was to reduce the amount of effort of an analyst that is required to correctly detect anomalies. We described several methods for computing SFEs and introduced a new framework that allows for large-scale quantitative evaluation of explanation methods. Our experiments indicated that, overall, the Sequential Marginal method for computing SFEs is the preferred method among those introduced in this paper.


%
%
%

\end{document}